\documentclass[conference]{IEEEtran}
\IEEEoverridecommandlockouts
\usepackage{cite}
\usepackage{amsmath,amssymb,amsfonts}
\usepackage{algorithmic}
\usepackage{graphicx}
\usepackage{textcomp}
\usepackage{xcolor}
\def\BibTeX{{\rm B\kern-.05em{\sc i\kern-.025em b}\kern-.08em
    T\kern-.1667em\lower.7ex\hbox{E}\kern-.125emX}}
\begin{document}

\title{Privacy Preserving Properties \\of \\Vision Classifiers\\
}


\author{\IEEEauthorblockN{1\textsuperscript{st} Pirzada Suhail}
\IEEEauthorblockA{\textit{IIT Bombay} \\
Mumbai, India \\
psuhail@iitb.ac.in}
\and
\IEEEauthorblockN{2\textsuperscript{nd} Amit Sethi}
\IEEEauthorblockA{\textit{IIT Bombay} \\
Mumbai, India \\
asethi@iitb.ac.in}
}

\maketitle

\begin{abstract}

Vision classifiers are often trained on proprietary datasets containing sensitive information, yet the models themselves are frequently shared openly under the privacy-preserving assumption. Although these models are assumed to protect sensitive information in their training data, the extent to which this assumption holds for different architectures remains unexplored. This assumption is challenged by inversion attacks which attempt to reconstruct training data from model weights, exposing significant privacy vulnerabilities. In this study, we systematically evaluate the privacy-preserving properties of vision classifiers across diverse architectures, including Multi-Layer Perceptrons (MLPs), Convolutional Neural Networks (CNNs), and Vision Transformers (ViTs). Using network inversion-based reconstruction techniques, we assess the extent to which these architectures memorize and reveal training data, quantifying the relative ease of reconstruction across models. Our analysis highlights how architectural differences, such as input representation, feature extraction mechanisms, and weight structures, influence privacy risks. By comparing these architectures, we identify which are more resilient to inversion attacks and examine the trade-offs between model performance and privacy preservation, contributing to the development of secure and privacy-respecting machine learning models for sensitive applications. Our findings provide actionable insights into the design of secure and privacy-aware machine learning systems, emphasizing the importance of evaluating architectural decisions in sensitive applications involving proprietary or personal data.

\end{abstract}


\section{Introduction}

The advent of modern vision classifiers has revolutionized a wide range of applications, from autonomous vehicles and medical imaging to facial recognition. These models, often trained on proprietary datasets, have become a cornerstone of advancements in artificial intelligence (AI). However, the growing practice of sharing pre-trained models has raised critical concerns about the privacy of the training data. Many assume that the act of sharing a trained model inherently preserves the privacy of the underlying dataset, but this assumption is increasingly being challenged by research demonstrating the potential for data leakage. The question remains: to what extent can different model architectures safeguard sensitive information against reconstruction or inversion attacks?

Model inversion attacks, where an adversary attempts to reconstruct the training data from a model's weights or output, highlight the vulnerability of machine learning systems to privacy breaches. This becomes particularly alarming in domains like healthcare, where proprietary datasets often contain highly sensitive information, or in applications involving biometric data, where privacy is paramount. Vision classifiers, in particular, process inherently personal and identifiable information, such as faces or medical scans, making it imperative to evaluate their privacy-preserving properties comprehensively.

In this study, we systematically assess the privacy-preserving capabilities of three prominent architectures: Multi-Layer Perceptrons (MLPs), Convolutional Neural Networks (CNNs)\cite{dosovitskiy2021imageworth16x16words}, and Vision Transformers (ViTs)\cite{oshea2015introductionconvolutionalneuralnetworks}. These architectures differ significantly in their structure, feature extraction mechanisms, and input processing pipelines, which could influence their tendencies to memorize and inadvertently leak training data. For example, CNNs are designed to capture spatial hierarchies and local patterns, while ViTs, on the other hand, employ self-attention mechanisms that focus on global relationships between input elements, potentially resulting in different privacy implications. By comparing these architectures, we aim to understand their relative vulnerabilities and the factors that contribute to privacy risks.

To evaluate the privacy-preserving properties of these classifiers, we utilize network inversion-based reconstruction techniques as in \cite{suhail2024net}. These methods attempt to recover the training data entirely from the model weights, providing a quantifiable measure of privacy leakage. Such reconstruction attacks exploit the tendency of models to memorize specific details about their training data, particularly when the training set is small or when over-parameterized architectures are used. By systematically applying these techniques, we assess the relative ease of reconstruction across architectures and analyze the impact of architectural differences on privacy.

We conduct our analysis in the most extreme case of privacy risk: at the end of the training process, without any knowledge of the training process itself, without utilizing any unobvious prior information, without any auxiliary datasets and relying on a single trained model. This setup provides a stringent evaluation of privacy risks, focusing on the inherent vulnerabilities of the model architectures. Further the quality of the reconstructed samples is assessed by comparing them to the original training samples using a similarity metric. Reconstructions with higher Structural Similarity Index Measure (SSIM) values indicate a greater privacy risk, as they suggest more effective memorization of the training data by the model. 

Our findings highlight that architectural differences in processing input images, feature extraction, and weight structures contribute to varying degrees of privacy leakage. We apply our evaluation to multiple benchmark datasets, including MNIST, FashionMNIST, CIFAR-10, and SVHN that cover a wide range of complexities and image types, allowing us to study how different architectures behave under varying data conditions.

\section{Related Works}

Privacy concerns in machine learning have led to extensive research in Privacy-Preserving Machine Learning (PPML), particularly in mitigating risks related to membership inference, attribute inference, and model inversion attacks. A foundational study by \cite{8677282} provides an overview of privacy threats in ML, including model inversion, and explores defenses such as differential privacy, homomorphic encryption, and federated learning. Similarly, \cite{xu2021privacypreservingmachinelearningmethods} introduces the Phase, Guarantee, and Utility (PGU) triad, a framework to evaluate PPML techniques across different phases of the ML pipeline. These studies highlight the need for privacy-preserving methods but primarily focus on algorithmic-level defenses rather than evaluating inherent vulnerabilities in different model architectures. Unlike these approaches, our study investigates the privacy risks posed by architectural design choices by analyzing how MLPs, CNNs, and ViTs differ in their susceptibility to model inversion attacks.

Network inversion has emerged as a powerful technique to understand how neural networks encode and manipulate training data. Initially developed for interpretability \cite{KINDERMANN1990277,784232}, it has since been shown to reconstruct sensitive training samples, raising significant privacy concerns \cite{Wong2017NeuralNI,ad}. Early works on network inversion focused on fully connected networks (MLPs) \cite{KINDERMANN1990277,SAAD200778}, demonstrating that they tend to memorize training data, making them vulnerable to inversion attacks. Evolutionary inversion procedures \cite{784232} improved the ability to capture input-output relationships, providing deeper insights into model memorization behavior. More recent studies extended these inversion techniques to CNNs \cite{ad}, showing that hierarchical feature extraction does not necessarily prevent training data leakage. The introduction of ViTs has further complicated this issue, as their global self-attention mechanisms process data differently than CNNs, raising new questions about how they store training information and whether their memorization patterns lead to higher or lower inversion risks.

In adversarial settings, model inversion attacks aim to reconstruct sensitive data by exploiting a model's predictions, gradients, or weights \cite{ad,kumar2019modelinversionnetworksmodelbased}. These attacks have been shown to succeed even without direct access to the training process, as demonstrated by \cite{9833677}, where an adversary with auxiliary knowledge reconstructs sensitive samples. Gradient-based inversion attacks further exacerbate these risks by leaking sensitive training information through shared gradients in federated learning setups \cite{pmlr-v206-wang23g}.

To improve the stability of inversion processes, recent works have explored novel optimization techniques. For example, \cite{liu2022landscapelearningneuralnetwork} proposed learning a loss landscape to make gradient-based inversion faster and more stable. Alternative approaches, such as encoding networks into Conjunctive Normal Form (CNF) and solving them using SAT solvers, offer deterministic solutions for inversion, as introduced by \cite{suhail2024network}. Although computationally expensive, these methods ensure diversity in the reconstructed samples by avoiding shortcuts in the optimization process.

Model Inversion (MI) attacks have also been extended to scenarios involving ensemble techniques, where multiple models trained on shared subjects or entities are used to guide the reconstruction process. The concept of ensemble inversion, as proposed by \cite{wang2021reconstructingtrainingdatadiverse}, enhances the quality of reconstructed data by leveraging the diversity of perspectives provided by multiple models. By incorporating auxiliary datasets similar to the presumed training data, this approach achieves high-quality reconstructions with sharper predictions and higher activations. This work highlights the risks posed by adversaries exploiting shared data entities across models, emphasizing the importance of robust defense mechanisms.

Reconstruction methods for training data have evolved significantly, focusing on improving the efficiency and accuracy of recovering data from models. Traditional optimization-based approaches relied on iteratively refining input data to match a model’s outputs or activations \cite{Wong2017NeuralNI}. More recent advancements have leveraged generative models, such as GANs and autoencoders, to synthesize high-quality reconstructions. These techniques aim to approximate the distribution of training data while maintaining computational efficiency. In the context of privacy risks, works like \cite{haim2022reconstructingtrainingdatatrained} demonstrated that significant portions of training data could be reconstructed from neural network parameters in binary classification settings. This work was later extended to multi-class classification by \cite{buzaglo2023reconstructingtrainingdatamulticlass}, showing that higher-quality reconstructions are possible and revealing the impact of regularization techniques, such as weight decay, on memorization behavior.

The ability to reconstruct training data from model gradients also presents a critical privacy challenge. The study by \cite{wang2023reconstructingtrainingdatamodel} demonstrated that training samples could be fully reconstructed from a single gradient query, even without explicit training or memorization. Recent advancements like \cite{oz2024reconstructingtrainingdatareal}, adapt reconstruction schemes to operate in the embedding space of large pre-trained models, such as DINO-ViT and CLIP. This approach, which introduces clustering-based methods to identify high-quality reconstructions from numerous candidates, represents a significant improvement over earlier techniques that required access to the original dataset. While \cite{pmlr-v162-guo22c} extended differential privacy guarantees to training data reconstruction attacks.

In this paper, we build upon prior work on network inversion and training data reconstruction. Drawing from works like \cite{suhail2024networkcnn} and \cite{suhail2024networkinversionapplications}, we employ network inversion methods to understand the internal representations of neural networks and the patterns they memorize during training. Our study systematically compares the privacy-preserving properties of different vision classifier architectures, including MLPs, CNNs, and ViTs, using network inversion-based reconstruction techniques \cite{suhail2024net}. By evaluating these techniques across datasets such as MNIST, FashionMNIST, CIFAR-10, and SVHN, we explore the impact of architectural differences, input processing mechanisms, and weight structures on the susceptibility of models to inversion attacks.

\section{Methodology}

\subsection{Overview}
In this study, we investigate the ease of reconstruction and the extent of memorization in trained vision classifiers based on different architectures. Our primary focus is on analyzing the most extreme case of training data reconstruction, where the inversion process relies almost entirely on the input-output relationships of the trained model and its learned weights. Unlike prior reconstruction approaches that leverage pre-trained models, auxiliary datasets, gradient information from the training process, or other unobvious priors, our method seeks to reconstruct training data with minimal external dependencies. This approach allows us to systematically evaluate how different architectures—Multi-Layer Perceptrons (MLPs), Convolutional Neural Networks (CNNs), and Vision Transformers (ViTs)—differ in their ability to preserve or expose sensitive training data.

To perform network inversion and data reconstruction, we build upon the methodology introduced in \cite{suhail2024networkcnn, suhail2024networkinversionapplications}, which has primarily focused on CNN-based classifiers. We extend this approach to other architectures, particularly MLPs and ViTs, to assess their relative vulnerability to inversion attacks. Briefly, network inversion techniques aim to generate inputs that align with the learned decision boundaries of a classifier by training a conditioned generator to reconstruct data that maximally activates specific output neurons. In its standard form, this inversion process does not necessarily yield images resembling actual training samples but instead produces arbitrary inputs that satisfy the model’s learned function. However, by modifying the inversion procedure following \cite{suhail2024network, suhail2024net}, we incentivize the generator to reconstruct training-like data by leveraging key properties of the classifier with respect to its training data. These modifications allow us to better quantify the extent of memorization across different architectures and assess the associated privacy risks. The proposed approach to Network Inversion and subsequent training data reconstruction uses a carefully conditioned generator that learns the data distributions in the input space of the trained classifier.

\subsection{Classifier Architectures}
We perform inversion and reconstruction on classifiers based on three distinct architectures:

\begin{itemize}
    \item \textbf{Multi-Layer Perceptrons (MLPs)}: These are fully connected networks where each neuron in one layer is connected to every neuron in the next layer. MLPs process flattened input images, lacking any inherent spatial hierarchy. The inversion and subsequent reconstruction will be performed using the logits and penultimate fully connected layers.
    \item \textbf{Convolutional Neural Networks (CNNs)}: CNNs use convolutional layers to extract hierarchical spatial features from images, enabling them to effectively capture local patterns. In this case the features from the fully connected layers are used after flattening the output of convolutional layers along with the logits in the last layer to perform inversion.
    \item \textbf{Vision Transformers (ViTs)}: ViTs utilize self-attention mechanisms to capture global dependencies across an image. This architecture is particularly effective in modeling long-range interactions within an image. In ViTs we particularly look at the classification token embeddings and use it in the same way as above.
\end{itemize}

These architectures have inherently different memory capacities and generalization properties, affecting their susceptibility to reconstruction attacks.

\subsection{Vector-Matrix Conditioned Generator}
The generator in our approach is conditioned on vectors and matrices to ensure that it learns diverse representations of the data distribution. Unlike simple label conditioning, the vector-matrix conditioning mechanism encodes the label information more intricately, allowing the generator to better capture the input space of the classifier. 

The generator is initially conditioned using $N$-dimensional vectors for an $N$-class classification task. These vectors are derived from a normal distribution and are softmaxed to form a probability distribution. They implicitly encode the labels, promoting diversity in the generated images.

Further, a Hot Conditioning Matrix of size $N \times N$ is used for deeper conditioning. In this matrix, all elements in a specific row or column are set to $1$, corresponding to the encoded label, while the rest are $0$. This conditioning is applied during intermediate stages of the generation process to refine the diversity of the outputs.

\subsection{Training Data Properties}
The classifier exhibits specific properties when interacting with training data, which are exploited to facilitate the reconstruction of training-like samples:
\begin{figure*}[t]
\centering
\includegraphics[width=1\textwidth]{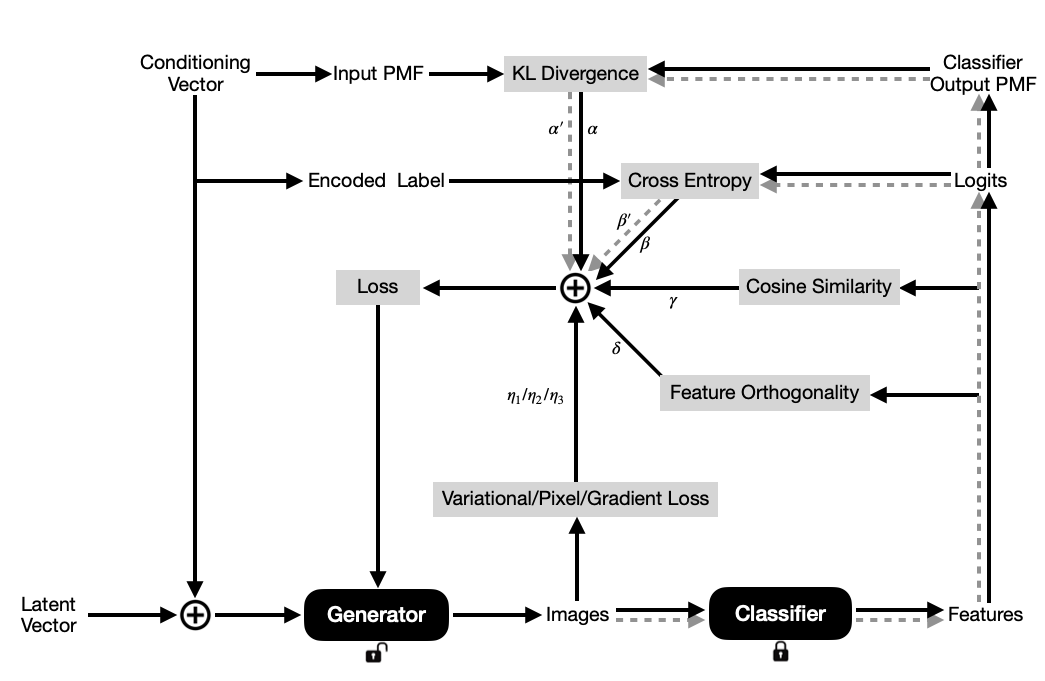} 
\caption{Schematic Approach to Training-Like Data Reconstruction using Network Inversion}
\label{fig:reconstruction}
\end{figure*}

\begin{itemize}
    \item \textbf{Model Confidence:} The classifier is more confident when predicting labels for training samples compared to random samples. Hence, in order to take this into account we condition the generation on one-hot vectors and then minimise its KL Divergence from the classifier's output enforcing generation of samples that are confidently classified buy the classifier. This can be expressed as:
    \begin{equation}
    P(y_{\text{in}} | x_{\text{in}}; \theta) \gg P(y_{\text{ood}} | x_{\text{ood}}; \theta)
    \end{equation}

    \item \textbf{Robustness to Perturbations:} During training the model gets to observe a diverse set of data including different variations of the images in the same class. Due to which the model is relatively robust to perturbations around training data compared to random inverted samples, meaning small changes do not significantly affect predictions. We take this into account in by perturbing the generated images and then ensuring that the perturbed images also produce similar output,
    \begin{equation}
    \frac{\partial f_{\theta}(x_{\text{in}})}{\partial x_{\text{in}}} \ll \frac{\partial f_{\theta}(x_{\text{ood}})}{\partial x_{\text{ood}}}
    \end{equation}

    \item \textbf{Gradient Behavior:} By virtue of training the model on a certain dataset, the gradient of the loss with respect to model weights is expected to be lower for training data compared to random inverted samples, as the model has already been optimized on it, Hence we add a penalty on the gradients that encourages the generation of samples with low gradients as:
    \begin{equation}
    \|\nabla_{\theta} L(f_{\theta}(x_{\text{in}}), y_{\text{in}})\| \ll \|\nabla_{\theta} L(f_{\theta}(x_{\text{ood}}), y_{\text{ood}})\|
    \end{equation}
\end{itemize}

\subsection{Training Data Reconstruction}
The reconstruction process utilizes the generator to produce training-like samples by taking into account the specific properties of the training data with respect to the classifier. The approach is schematically illustrated in Figure \ref{fig:reconstruction}.

The primary loss function used for reconstruction, 
\begin{align*}
\mathcal{L}_{\text{Recon}} = & \; \alpha \cdot \mathcal{L}_{\text{KL}} 
+ \alpha' \cdot \mathcal{L}_{\text{KL}}^{\text{pert}}
+ \beta \cdot \mathcal{L}_{\text{CE}} 
+ \beta' \cdot \mathcal{L}_{\text{CE}}^{\text{pert}} \\
& + \gamma \cdot \mathcal{L}_{\text{Cosine}} 
+ \delta \cdot \mathcal{L}_{\text{Ortho}} \\
& + \eta_1 \cdot \mathcal{L}_{\text{Var}} 
+ \eta_2 \cdot \mathcal{L}_{\text{Pix}} 
+ \eta_3 \cdot \mathcal{L}_{\text{Grad}}
\end{align*}

where \( \mathcal{L}_{\text{KL}} \) is the KL Divergence loss, \( \mathcal{L}_{\text{CE}} \) is the Cross Entropy loss, \( \mathcal{L}_{\text{Cosine}} \) is the Cosine Similarity loss, and \( \mathcal{L}_{\text{Ortho}} \) is the Feature Orthogonality loss. The hyperparameters \( \alpha, \beta, \gamma, \delta \) control the contribution of each individual loss term defined as:
\[
\mathcal{L}_{\text{KL}} = D_{\text{KL}}(P \| Q) = \sum_{i} P(i) \log \frac{P(i)}{Q(i)}
\]
\[
\mathcal{L}_{\text{CE}} = -\sum_{i} y_{i} \log(\hat{y}_{i})
\]
\[
\mathcal{L}_{\text{Cosine}} = \frac{1}{N(N-1)} \sum_{i \neq j} \cos(\theta_{ij})
\]
\[
\mathcal{L}_{\text{Ortho}} = \frac{1}{N^2} \sum_{i, j} (G_{ij} - \delta_{ij})^2
\]
where \( D_{\text{KL}} \) represents the KL Divergence between the input distribution \( P \) and the output distribution \( Q \), \( y_{i} \) is the set encoded label, \( \hat{y}_{i} \) is the predicted label from the classifier, \( \cos(\theta_{ij}) \) represents the cosine similarity between features of generated images \( i \) and \( j \), \( G_{ij} \) is the element of the Gram matrix, and \( \delta_{ij} \) is the Kronecker delta function. \( N \) is the number of feature vectors in the batch.

Further to take reconstruction into account we also use \(\mathcal{L}_{\text{KL}}^{\text{pert}}\) and \(\mathcal{L}_{\text{CE}}^{\text{pert}}\) that represent the KL divergence and cross-entropy losses applied on perturbed images, weighted by \( \alpha'\) and  \(\beta' \)respectively while \(\mathcal{L}_{\text{Var}}\), \(\mathcal{L}_{\text{Pix}}\) and \(\mathcal{L}_{\text{Grad}}\) represent the variational loss, Pixel Loss and penalty on gradient norm each weighted by \( \eta_1\), \( \eta_2\), and \(\eta_3\) respectively and defined for an Image \(I\) as:
\begin{align*}
\mathcal{L}_{\text{Var}} = \frac{1}{N} \sum_{i=1}^{N} \Bigg( \sum_{h,w} & \Big( ( I_{i, h+1, w} - I_{i, h, w} )^2 \\
& + ( I_{i, h, w+1} - I_{i, h, w} )^2 \Big) \Bigg)
\end{align*}
\[
\mathcal{L}_{\text{Grad}} = \left\| \nabla_{\theta} L(f_{\theta}(I), y) \right\|
\]

\[
\mathcal{L}_{\text{Pix}} = \sum \max(0, -I) + \sum \max(0, I - 1)
\]

By integrating these loss components, the generator is trained to produce samples that closely resemble the training data, thus revealing the extent of memorization within the classifier.

In this section, we present the experimental results obtained by applying the reconstruction technique on the MNIST \cite{deng2012mnist}, FashionMNIST \cite{xiao2017fashionmnistnovelimagedataset}, SVHN, and CIFAR-10 \cite{cf} datasets. Our goal is to evaluate the ease of reconstruction and the extent of memorization in different vision classifier architectures by training a generator to produce images that resemble the training data. The classifier is first trained normally on a given dataset and then held in evaluation mode for the purpose of reconstruction. The conditioned generator, which takes as input latent vectors and conditioning information, is trained to generate images that the classifier maps to specific labels.

We evaluate three vision classifier architectures: Multi-Layer Perceptrons (MLPs), Convolutional Neural Networks (CNNs), and Vision Transformers (ViTs). We implement a 5-layer MLP with Batch Normalization, Leaky ReLU activations, and Dropout layers \cite{JMLR:v15:srivastava14a} to mitigate memorization tendencies. The CNN consists of three convolutional layers followed by batch normalization \cite{pmlr-v37-ioffe15}, dropout layers, and a fully connected layer for classification. The ViT classifier uses a transformer-based architecture with three self-attention layers, each containing four attention heads while the input images are divided into non-overlapping patches of size \(4 \times 4\).

The generator, instead of traditional label embeddings, follows a Vector-Matrix Conditioning approach to ensure diverse and structured image generation. The class labels are encoded into random softmaxed vectors concatenated with the latent vector, followed by multiple layers of transposed convolutions, batch normalization, and dropout layers to encourage diversity in generated images. Once the concatenated latent and conditioning vectors are upsampled to \(N \times N\) spatial dimensions for an \(N\)-class classification task, they are concatenated with a conditioning matrix for further generation up to the required image size (\(28 \times 28\) for MNIST and FashionMNIST, and \(32 \times 32\) for SVHN and CIFAR-10).

To assess the extent of memorization and the relative ease of reconstruction across different architectures, we use the Structural Similarity Index Measure (SSIM). The goal was to evaluate how different architectures handle privacy risks by comparing the quality of reconstructed images. SSIM quantifies the perceptual similarity between two images based on luminance, contrast, and structural similarity components. In this study, SSIM is used to compare the reconstructed images with their closest matches from the training dataset. Higher SSIM values indicate stronger resemblance to training samples, suggesting a greater privacy risk due to increased memorization by the classifier.

Table~\ref{tab:ssim_values} presents the SSIM scores between reconstructed samples and training data, averaged across all classes for each dataset and architecture. Higher SSIM values suggest greater memorization and weaker privacy preservation, as the reconstructed samples strongly resemble real training data.

\begin{table}[h]
\centering
\caption{SSIM values for reconstructed samples across different architectures and datasets.}
\label{tab:ssim_values}
\resizebox{\linewidth}{!}{
\begin{tabular}{|l|c|c|c|}
\hline
\textbf{Dataset}       & \textbf{MLP} & \textbf{ViT} & \textbf{CNN} \\ \hline
\textbf{MNIST}         & 0.83         & 0.78         & 0.73         \\ \hline
\textbf{FashionMNIST}  & 0.74         & 0.64         & 0.63         \\ \hline
\textbf{SVHN}          & 0.71         & 0.68         & 0.69         \\ \hline
\textbf{CIFAR-10}      & 0.65         & 0.62         & 0.58         \\ \hline
\end{tabular}
}
\end{table}

From Table~\ref{tab:ssim_values}, we observe that MLPs exhibit the highest SSIM scores across all datasets, suggesting that they retain more training data details than CNNs and ViTs. This is likely due to their fully connected nature and lack of spatial inductive biases, leading to higher memorization tendencies. Also individual pixels in the images have dedicated weights associated, making memorization easier.

ViTs generally show lower SSIM values than MLPs but remain slightly higher than CNNs, indicating that self-attention mechanisms contribute to retaining finer details in the reconstructed images. Unlike CNNs, ViTs lack pooling layers, which means that more spatial information is preserved, making inversion attacks more feasible. 

CNNs show the lowest SSIM values across all datasets, suggesting that they inherently discard more specific input details due to weight sharing, local receptive fields, and pooling operations. This abstraction reduces direct memorization and makes CNNs relatively more privacy-preserving compared to MLPs and ViTs.

\begin{figure}[ht]
\centering
\includegraphics[width=0.95\linewidth]{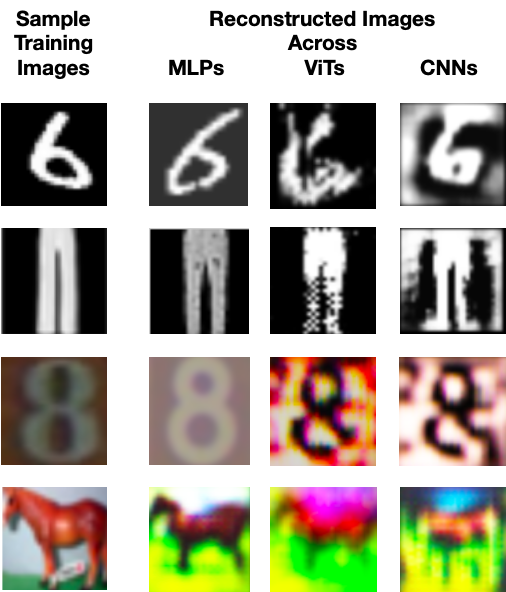} 
\caption{Comparison of reconstructed samples across MLP, ViT, and CNN architectures. The first column represents actual training samples, while subsequent columns show corresponding reconstructed images.}
\label{fig:sam}
\end{figure}

Figure~\ref{fig:sam} presents a qualitative comparison of reconstructed samples across different classifier architectures. The first column depicts actual training samples, while the subsequent columns display reconstructed images generated from MLPs, ViTs, and CNNs. Notably, MLP-generated reconstructions appear most similar to the original samples, while CNNs yield more abstract representations, indicating lower memorization.

These results highlight the variation in privacy risks across architectures, with MLPs being the most susceptible to memorization, followed by ViTs, and CNNs exhibiting the lowest reconstruction fidelity. The findings emphasize the importance of architectural choices in designing privacy-aware models, where CNNs may be preferable for applications requiring privacy preservation.

\section{Conclusion and Future Work}
In this paper, we systematically evaluated the privacy-preserving properties of vision classifiers by analyzing the extent of memorization and the ease of training data reconstruction across different architectures. Our experimental results, quantified through SSIM scores, indicate that MLPs tend to memorize more information about training samples compared to CNNs and ViTs, making them more prone to reconstruction. These findings highlight the crucial role of architectural choices in mitigating privacy risks, emphasizing the need for privacy-aware model deployment, especially in sensitive applications.

Future research can extend this study by exploring additional factors that influence privacy leakage, such as model depth, dataset complexity, and the impact of regularization techniques. In the case of ViTs it would be also of interest to see how the reconstructions differ with varying patch size. Investigating the impact of differential privacy mechanisms in reducing reconstruction risks could provide valuable insights into enhancing privacy.

\bibliographystyle{IEEEtran}  
\bibliography{references} 

\end{document}